\documentclass[letterpaper, 10 pt, conference]{ieeeconf}  

\IEEEoverridecommandlockouts                              

\usepackage{amsmath} 
\usepackage{xcolor}
\usepackage{soul}
\usepackage{comment}

\title{\LARGE \bf
Data-Driven Hierarchical Open Set Recognition
}

\author{Andrew Hannum$^{1}$, Max Conway$^{2}$, Mario Lopez$^{3}$, André Harrison$^{4}$
\thanks{*This work was not supported by any organization}
\thanks{$^{1}$Andrew Hannum is with Faculty of Ritchie School of Engineering and Computer Science,
        University of Denver, 2155 E Wesley Ave, Denver, CO 80208
        {\tt\small andrew.hannum@du.edu}}%
\thanks{$^{2}$Max Conway with the Department of Computer Science, University of Colorado Boulder,
        430 UCB, 1111 Engineering Dr, Boulder, CO 80309
        Boulder, CO 45435, USA
        {\tt\small max.conway@colorado.edu}}%
\thanks{$^{3}$Mario Lopez is with Faculty of Ritchie School of Engineering and Computer Science,
        University of Denver, 2155 E Wesley Ave, Denver, CO 80208
        {\tt\small mario.lopez@du.edu}}%
\thanks{$^{4}$André Harrison is with DEVCOM Army Research Laboratory,
        2800 Powder Mill Rd, Adelphi, MD 20783
        {\tt\small andre.v.harrison2.civ@army.mil}}%
}

\begin{document}

\maketitle
\thispagestyle{empty}
\pagestyle{empty}

As robots gain greater ubiquity and robustness they are expected to operate in more than just factory environments with constant environmental settings, but in the dynamic and ever-changing real world. However, the open nature of the world poses a fundamental problem in robotics and computer vision. For instance, a warehouse robot trained to navigate around workers and move boxes may encounter entities it was not trained to recognize when used outside of the factory environment, which can lead to task failure. To achieve robust perception, open set recognition, which identifies when an input belongs to an unknown class not included in training, has gained significant attention in both fields.

\section{Background and Motivation}

One approach to handle unknown instances is to use a hierarchical structure of classes. This allows the model to not only distinguish an unknown class from the set of know classes but is also able to describe a new class relative to known classes, even when precise classification is not possible \cite{wilt2021creating, bennette2021hierarchical, samplawski2019integrating}. For example, a self-driving car might classify an ambiguous input as a cyclist, covering both motorcyclists and bicyclists, without needing to distinguish between the two. The major challenge with hierarchical approaches is the effort required to create the hierarchies \cite{wilt2021creating, bennette2021hierarchical}, which can encode human biases. For example, in Linnaeus' Hierarchical Categorization system, a dolphin belongs to the class Mammalia (mammals) while a shark belong to the class Chondrichthyes (fish), placing a dolphin closer to a dog than to a shark, which may be counter intuitive in a visual classification system.

\section{Proposed Approach}

We propose a novel procedure for open set recognition that provides additional information by relating an unknown class sample to known classes, without the need for manually obtained relational information. We first build a hierarchy of known classes in an embedding space using constrained agglomerative clustering, a hierarchical clustering method that merges data points or clusters while respecting predefined constraints on which items can be grouped together. New samples are then classified by finding the best matching node in this hierarchy. We demonstrate our approach on the Animals with Attributes 2 (AwA2) dataset \cite{AnimalsWithAttributes2}. Preliminary results obtain an AUC ROC score of 0.82 and an average utility score of 0.85, which is comparable to state-of-the-art open set methods that only distinguish between known and unknown classes without further clarification \cite{miller2024open}, \cite{bennette2021hierarchical}, \cite{wang_openauc_2023}, \cite{vaze_open-set_2022}, \cite{cevikalp_deep_2021}, \cite{neal_open_2018}. However, the benefit of our method lies not only in providing additional relational information, but in doing so through an automatically generated hierarchy from the training data, rather than relying on manually created hierarchies. We believe that this approach can be useful in various practical settings, e.g., robotics and computer vision, where unknown classes are common, and in low-resource settings where auxiliary information is scarce, difficult to produce, or introduces undesirable biases or associations.

Our main contributions are the proposal of this data-driven hierarchical OSR approach and the introduction of a new metric, \emph{Concentration Centrality} (CC). This metric is a measure of how well-concentrated the predicted labels are in the hierarchy tree, for samples belonging to a given class. It can be used to give a measure of the overall consistency of a hierarchical classification model.

Training of an open set model consists of a split of the dataset into training and testing sets, reserving some classes only for testing as unseen classes ($U$). We create a hierarchy ($T$) using the training dataset, which contains samples of known classes ($C$) using existing agglomerative clustering techniques: embeddings of known classes are obtained from an encoder ($E$), and the resulting class embeddings are used as inputs into a constrained agglomerative clustering algorithm, which merges classes with the minimum distance defined by a distance function $D(c_i, c_j) \mid c_i, c_j \in C$.

Given the hierarchy $T$, we classify a sample $x \in C \cup U$ as belonging to a node in the hierarchy $t \in T$. If $x$ belongs to a known class $c_i \in C$, it should be classified to a leaf node representing that class. If $x$ belongs to an unknown class, it should be classified as an internal node.

We propose two approaches for classification: \emph{score-based} and \emph{traversal-based}. In score-based classification, a function $S(x,t) \mid t \in T$ gives a value representing the fit quality for a sample and node, the node with the best fit being the final classification. In traversal-based classification, we determine if the sample best fits the root, left child, or right child, using a combination of traditional binary classifier models and outlier detection models, trained individually on every node of the tree (in the case of outlier detection) and the children of every internal node (in the case of binary classifiers). Traversal continues recursively until reaching either an intermediate node where the sample is classified as inlier, or a leaf node.

\section{Evaluation Metrics}

To evaluate the model's performance, we use various metrics including common measures of accuracy such as AUC ROC, F1 Score, Precision vs Recall, etc. In addition, for known classes, it is preferable for the model to identify lower ancestors in the hierarchy. We use the utility metric from \cite{samplawski2019integrating} to measure prediction quality:
\[
\displaystyle \emph{Utility}(C, M) = \frac{\sum_{i=0}^{n} \frac{depth(M(x_i))}{depth(c_i)}}{n}
\]

Finally, for unseen classes, we are interested in the consistency of their placement in the hierarchy. We use our new metric, the \emph{Class Concentration Centrality} (CCC), to measure this consistency.

Given \( P(t, k) \), the proportion of class \( k \) assignments at node \( t \). The \emph{Concentration Centrality} (CC) for a class \( k \) at a node \( t \) is:
\[
\mbox{CC}(t, k) = 1 - \sum_{t' \in T} \frac{d(t, t')}{d_{\max}(t)} P(t', k)
\]
where \( d(t, t') \) is the shortest path distance between nodes \( t \) and \( t' \) in the undirected version of \( T \), and \( d_{\max}(t) \) is the maximum distance from \( t \) to any other node in \( T \).

The proposed CC measure shares similarities with the existing \emph{closeness centrality} graph metric \cite{bavelas_communication_1950}. Both are measures of how ``close" a given node is to other nodes in a graph, using shortest path distances and favoring nodes with shorter paths to others. For closeness centrality the measure is calculated by the reciprocal of the sum of shortest path distances from a node to all other nodes in the network.

CC differs by incorporating class-based weighting on the shortest path calculations, normalizing by maximum distance from the focal node, and scaling to \([0,1]\). It measures class concentration around a node by combining normalized distances with class proportions — essentially a class-weighted, normalized closeness centrality.

The node \( n^\star \) with the maximum CC score is considered the optimal concentration center for class \( k \), giving the CCC score for a given class:
\[
n^\star = \arg\max_{n \in N} CC(n, k)
\]
\[
\mbox{CCC}(k) = \mbox{CC}(n^\star, k)
\]

The overall consistency of the model on unseen classes can be measured by the average CCC score across all classes.

Our results on the AwA2 dataset were comparable to existing state-of-the-art methods while providing additional information on unknown classes. We observed that instances of unknown classes had highly concentrated placements in the hierarchy, resulting in high CCC scores. Instances of known classes were placed close to, or in, the leaves for the known class, leading to high utility scores and high accuracy scores. These findings validate the effectiveness of automated methods for hierarchy generation and provide new measurements for consistency of placement in the hierarchy, contributing meaningfully to hierarchical open set recognition.

In summary and conclusion, we propose a novel data-driven hierarchical open set recognition method that uses a constrained agglomerative clustering algorithm to build a hierarchy of known classes. We classify new samples by finding the best matching node in this hierarchy, or rejecting them if they do not fit any known class. The best models trained so far obtain an AUC ROC score of 0.82 and an average utility of 0.85. While this level of accuracy is not better than existing methods, it has the benefit of not requiring additional information beyond the typical supervised model requirements (i.e. labeled classes), which is advantageous for data-driven applications. Future work will focus on improving the accuracy and further validating the Concentration Centrality metric. We plan to run these experiments following the Large-Scale Open-Set Classification Protocols for ImageNet \cite{palechor2023large}. To our knowledge we will be the first hierarchical open set recognition paper to do so.

\bibliographystyle{unsrt}
\bibliography{references}

\end{document}